\documentclass[sigconf,natbib=true,square,numbers]{acmart}
\renewcommand\footnotetextcopyrightpermission[1]{} % removes footnote with conference information in first column

\usepackage{graphicx} % Required for inserting images
\usepackage{dsfont}
\usepackage{enumitem}
\usepackage{pgfplots}
\usepackage{subcaption}
\usepackage{booktabs}
\usepackage{float}
\usepackage{xspace}
\pgfplotsset{compat=newest}
\usepackage{multirow}
\usepackage[normalem]{ulem}
\usepackage{listings}
\usepackage[]{mdframed}
\usepackage[most]{tcolorbox}
\usepackage{alltt}
\usepackage{xspace}

\newcommand{\datamorgana}{DataMorgana\xspace}

\newcommand{\Ni}{({\em i})~}
\newcommand{\Nii}{({\em ii})~}
\newcommand{\Niii}{({\em iii})~}

\acmYear{2025}
\acmConference[LLM4Eval@WSDM '25 submission]{Submission to LLM4Eval @ WSDM}{Mar 10--14,
  2025}{Hannover, Germany}

\title{Generating Diverse Q\&A
Benchmarks for RAG Evaluation with \datamorgana}
\author{Simone Filice}
\email{filice.simone@gmail.com}
\affiliation{
  \institution{Technology Innovation Institute}
  \city{Haifa}
  \country{Israel}
}
\author{Guy Horowitz}
\email{guy.horowitz@tii.ae}
\affiliation{
  \institution{Technology Innovation Institute}
  \city{Haifa}
  \country{Israel}
}

\author{David Carmel}
\email{david.carmel@tii.ae}
\affiliation{
  \institution{Technology Innovation Institute}
  \city{Haifa}
  \country{Israel}
}

\author{Zohar Karnin}
\email{zohar.karnin@tii.ae}
\affiliation{
  \institution{Technology Innovation Institute}
  \city{Haifa}
  \country{Israel}
}
\author{Liane Lewin-Eytan}
\email{liane.lewineytan@tii.ae}
\affiliation{
  \institution{Technology Innovation Institute}
  \city{Haifa}
  \country{Israel}
}

\author{Yoelle Maarek}
\email{yoelle@yahoo.com}
\affiliation{
  \institution{Technology Innovation Institute}
  \city{Haifa}
  \country{Israel}
}
\begin{document}

\begin{abstract}
%Retrieval-Augmented Generation (RAG) systems combine Large Language Models (LLMs) with state-of-the-art information retrieval to address specialized and dynamic information needs.
Evaluating Retrieval-Augmented Generation (RAG) systems, especially in domain-specific contexts, requires benchmarks that address the distinctive requirements of the applicative scenario. Since real data can be hard to obtain, a common strategy is to use LLM-based methods to generate synthetic data. Existing solutions are general purpose: given a document, they generate a question to build a Q\&A pair. However, although the generated questions can be individually good, they are typically not diverse enough to reasonably cover the different ways real end-users can interact with the RAG system. 

We introduce here \datamorgana, a tool for generating highly customizable and diverse synthetic Q\&A benchmarks tailored to RAG applications. \datamorgana enables detailed configurations of user and question categories
%without being limited to pre-defined options, 
and provides control over their distribution within the benchmark. It uses a lightweight two-stage process, ensuring efficiency and fast iterations,
%ym redundant
while generating benchmarks that reflect the expected traffic.
%user types and needs, as well as varied question or query styles. 

We conduct a thorough line of experiments, showing quantitatively and qualitatively that \datamorgana surpasses existing tools and approaches in producing lexically, syntactically, and semantically diverse question sets across domain-specific and general-knowledge corpora. %\zk{the sentence after this seems redundant to me} %This approach enables generating benchmarks that align with specific requirements of RAG applications in various real-world scenarios, offering a significant advancement in benchmark generation.
%\lle{do we want to mention here also the challenge? or any specific result that stands out in the experiments?}\simo{Actually I would mention it here and not in the intro.}
\datamorgana will be made available to selected teams in the research community, as first beta testers, in the context of the upcoming SIGIR'2025 LiveRAG challenge to be announced in early February 2025.
\end{abstract}

\maketitle
\pagestyle{plain} % removes running headers

\section{Introduction}

\label{sec:intro}

Retrieval-Augmented Generation (RAG) \cite{lewis2020retrieval, gao2023retrieval} has recently gained a great deal of popularity, especially in specialized domains. It combines the strengths of Large Language Models (LLMs) with modern information retrieval by dynamically augmenting the LLM prompt with relevant information from an auxiliary corpus. This hybrid approach allows for more accurate and contextually relevant responses and thus mitigates LLMs limitations in handling specialized or frequently updated information.

Before adopting a RAG solution, however, it is critical to evaluate its effectiveness in the target environment, accounting not only for the environment's specific content (the RAG corpus) but also for its diverse types of users and their needs. Let us consider the typical RAG scenario of users asking questions over an enterprise-specialized corpus not memorized in the LLM. In order to evaluate the RAG solution, in the absence of a real question/query log, 
%or of a manually-generated Q\&A benchmark, 
the most common approach is to use an LLM to generate a Q\&A pair from a randomly selected document from the RAG corpus. %synthetic benchmark derived from the RAG corpus.
%Such benchmarks are typically assembled by prompting an LLM 
%(typically different from the one used within the RAG solution) 
%to generate questions about an input document sampled from the RAG corpus.
The major risk in applying this approach indiscriminately is that such synthetic benchmarks lack diversity and might not reflect the actual questions users would ask. 

%\lle{we contradict ourselves later, when saying this is not our target}\simo{Actually we want to have high coverage of all possible user behavior, so we want to mimic them, but for us it is more important to have enough test cases for each possible interaction type, than having a dataset that resembles the real traffic also in terms of interaction type frequencies} %ym removed to address Simone's point
%and suffer from a lack of lexical, syntactic and semantic diversity \simo{The concept of diversity is way more deep than lexical, syntactic and semantic, so I would not really mention them.}. It is important that generated questions cover various query types from simple factoid questions to complex analytical queries, and that they do not sound as if expressed by the same user. For instance in a medical RAG application, one would expect the benchmark to include questions that could be asked by patients as well as physicians and specialists with different semantic content and different language. 

% \zk{old paragraph:\\
% To this purpose, we propose here a new approach to generate synthetic benchmarks via a preliminary configuration stage that allows to specify the expected types of questions and users, and their expected distribution in the generated benchmark.
% This customization is especially needed for RAG applications in specialized domains, where the type of users and questions might highly vary.}

% \zk{new paragraph}
To this purpose, we propose here a new approach to generate synthetic benchmarks with two key properties.
{\em Straightforward and flexible customization}: Setting the way in 
which Q\&A pairs are generated is done via natural language descriptions, making customization accessible to non-technical users.
{\em Diverse generation}: For both end-users and questions we allow defining multiple categorizations, along with their distribution within the benchmark, without being restricted by pre-defined options. These categorizations are jointly used to get a combinatorial number of possibilities to define Q\&A pairs, leading to highly diverse benchmarks.

%ym agree - rephrased to remove ambiguity
%it as the RAG customer\simo{I think RAG customer will be easily confused with RAG end-user. What about RAG owner?} would know best what types of end-users will use the RAG solution and what types of requests they might ask. 

We have implemented this approach in a tool called \datamorgana, which we introduce in this paper.  \datamorgana %allows the generation of diverse and customized benchmarks via intuitive configuration capabilities that do not require any coding skill. %support customizing generated benchmarks in order to better mimic the behavior of a RAG solution's users. 
is designed to be lightweight and easily configurable, allowing for rapid experimentation with custom question and end-user categories. In this paper, we focus on the question generation capabilities of \datamorgana, to demonstrate via quantitative experiments that it supports higher diversity than related tools or approaches. 

Our key contributions in this work are as follows:
\begin{itemize}
    \item We introduce \datamorgana, a synthetic benchmarks generation tool, emphasizing easy customization and high diversity.
    % \item We introduce \datamorgana, a highly configurable tool for generating synthetic benchmarks, where any categories of questions and end-users can be easily defined along their distribution, unrestricted by pre-defined options, and without the need for reprogramming the system. %The configuration is done through a lightweight JSON-based process, eliminating the need for reprogramming the system.
    \item We guarantee the creation of high-coverage, highly diverse benchmarks, via a novel technique based on multiple end-user and question categorizations.
    % \item We guarantee the creation of a high-coverage, highly diverse benchmark, by allowing all combinations of end-user and question categories, thus achieving a combinatorial number of options.
    \item Through a comprehensive series of experiments on different corpora, we demonstrate the superiority of \datamorgana in achieving a higher diversity of generated questions compared to existing benchmark generation methods, across lexical, syntactic, and semantic dimensions.
\end{itemize}

%\ym{List here the key contributions after the entire paper is written including conclusion and future work}
%\simo{1) configurability is easy without coding to adapt to a desired scenario; 2) combinatorial number of prompts, 3) experimental results showing higher diversity}

\section{Related Work}
\label{sec:related}

\begin{figure*}[h]
\centering
\includegraphics[width=0.7\linewidth]{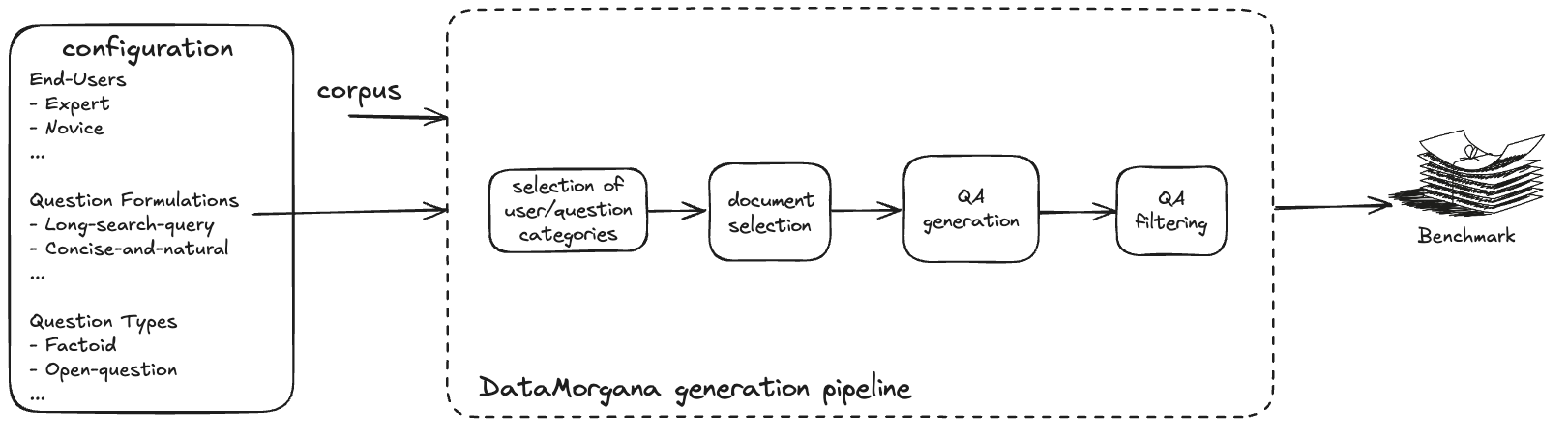}
\vspace{-0.4cm}
\caption{DataMorgana generation pipeline.}
\label{fig:datamorgana}
\end{figure*}

Recent advances in LLMs, with their tremendous zero-shot and few-shot generation capabilities,  have led to many research efforts in creating synthetic test benchmarks for question answering \cite{fei2022cqg,dong2023unified,yoon2023diversity, chen2024benchmarking, shakeri-etal-2020-end} and conversational dialog systems \cite{ling2020leveraging, do2022cohs}. 
Ideally, an optimal test set would comprise a large set of real user questions from a query log, paired with “golden answers” provided by experts.
In the absence of a perfect test set, we seek to generate questions similar to those asked by real users, along with answers inferred from a data source.
This task has received increasing interest in recent years from industrial and research communities because of its huge potential benefits in reducing the needed human labor in creating large-scale question-answer benchmarks. A comprehensive taxonomy of generation approaches can be found in \cite{zhang2021review,long2024llms}.

The common methodology for (question, answer) pairs generation is to follow the {\em generate then filter} paradigm. Given a corpus of documents, select at first a subset of documents; then, for each document leverage an LLM to generate some questions that can be answered by the given content. Next, ask the LLM to generate, for each of the questions, an answer, or a set of answers, based on the corresponding document. Finally, filter the generated (question, answer, document) tuples according to several criteria such as semantic similarity with golden questions, diversity, and more~\cite{yoon2023diversity}.

InPars~\citep{jeronymo2023inpars} follows this paradigm, focusing on question generation while skipping the answer generation process. Via few-shot examples, an LLM is induced to generate relevant questions for a given document. Then, each (question, document) pair is scored according to their inner similarity and only highly scored pairs are selected. 
Prompagator~\citep{dai2022promptagator}, and more recently ARES \cite{saad2024ares}, follow the same pipeline while keeping generated pairs only if the associated document appears on top of the result list when the question is submitted as a query to a given IR system.
\citet{shakeri-etal-2020-end} generate questions and answers from input documents using a fine-tuned encoder-decoder model and then filter them based on the model's perplexity score.
\citet{yuan2023selecting} proposed a prompt-based approach to selecting high-quality questions from a set of LLM-generated candidates. 

Uncontrolled generated content often tends to be monotonous and biased, hence limiting its applicability in downstream tasks~\cite{long2024llms}. The diversity of generated data is crucial for generating synthetic samples that mimic the diversified nature of real-world data, thereby preventing over-fitting and bias during model training or evaluation. 
\citet{yoon2023diversity} improve the diversity of generated questions by employing a recursive generation framework; they train a generative model (BART) to generate a question that differs from input questions, where the difference is measured with cosine similarity. At inference time previously generated questions are recursively fed back into the generation model to output different questions from reference questions. \citet{eo-etal-2023-towards} enhance Q\&A type diversity by training the Q\&A generator to cover various types of questions per document, based on interrogative question words (\{Who, When, What, Where, Why, How\}).

%\ym{I am missing a transition sentence here to introduce the three systems next, is there a way to characterize them?}
Recent studies have suggested enhancing administrative control over the types of generated questions.
In the {\em Know Your RAG} evaluation system~\cite{de2024know}, a taxonomy of question types is identified to cover different ways a user might interact with the system. The question generation process has three steps: \Ni a document is decomposed into statements, \Nii depending on the question type, new statements are generated based on the previously extracted ones; and \Niii one statement is selected as a base information to generate a question. Each of these steps is done via invoking an LLM.
%
% Some strategies were proposed to produce diverse questions from a collection of documents that better cover the desired question types. At first, by employing prompt engineering and multi-step LLM querying, then by fine-tuning small LLMs for this task. 
%
% \zk{Another relevant paper: \cite{eo2023towards} have the title "Towards Diverse and Effective Question-Answer Pair Generation from Children Storybooks"}
%

RAGAs~\cite{ragas} is a popular evaluation tool for RAG systems that additionally supports the generation of a synthetic Q\&A benchmark. At first, a knowledge graph (KG) is generated from the corpus by identifying entities, topics, and the relations between them; the test questions are then generated by an LLM based approach on the KG. Similarly to {\em Know Your RAG}, RAGAs considers different question types (single-hop vs multi-hop, specific versus abstract) as well as the user persona (senior, junior, etc.). This enriches the type of generated questions and improves diversity.
DeepEval~\cite{DeepEval} is another evaluation tool that supports generating a synthetic Q\&A benchmark. To encourage diversity they enable an evolutionary process where the generated questions are used towards generating new questions (i.e., evolved) according to 1 of 7 pre-defined evolution methods. 
In contrast,   \datamorgana, controls the diversity level with finer granularity via the customization of the question and user characteristics(See \S{\ref{sec:datamorgana}}). In \S{\ref{sec:assessing_diversity}}, we analyze the set of questions generated from the same corpus by a few of these systems, to compare their diversity with those generated by \datamorgana.

\section{DataMorgana}
\label{sec:datamorgana}

%\simo{Here an option we discussed is it present a simplified version of DataMorgana where we won't use conversations. The motivation is that this will \Ni simplify the dissertation, and \Nii simplify our experimental evaluation. Another option is justify the reduced experimental plan by the fact that existing solutions do not support conversations.}
%\dc{Maybe it would be better to describe DM with the conversation capability while in the experimental section focus on QA only.}
\datamorgana is designed to generate synthetic benchmarks for training and testing primarily RAG systems and possibly other systems that require Q\&A benchmarks. It differs from other tools by offering configuration capabilities that allow to easily generate benchmarks with high diversity.  It operates in two stages: a configuration stage, during which the \datamorgana admin user specifies their needs, and a generation stage during which \datamorgana leverages the input configuration to generate, with the assistance of an LLM, the desired benchmark.

\subsection{Configuration Stage}

%\zk{Need to add: (1) the fact that categorization have levels, ending up with a combinatorial number of types adding to diversity. (2) give the default question and end-user types (The default question categories are in Table~\ref{tab:question categories}). (3) provide some examples of different question and end-users for hypothetical scenarios (to demonstrate the value of allowing custom types).}

The configuration stage allows for the definition of detailed categorizations and associated categories for both questions and end-users, which provide high-level information on the expected traffic of the RAG application. There can be as many categorizations of questions and users as need be, as long as categories within a single categorization are mutually exclusive and each category is associated with its natural-language description and optionally its desired probability of distribution within the generated benchmark. 

The configuration is defined in a JSON file that includes all necessary information to customize the generated benchmark as desired. For instance, if a user wants to generate factoid and non-factoid ``experience'' questions, as defined in the six types (i.e., instructions, reason, evidence-based, comparison, experience, and debate) of non-factoid questions suggested in \cite{bolotova.2022}, they will include in their configuration file the following fragment. \\

%\lstinputlisting[basicstyle=\ttfamily\small,breaklines,showspaces=true,language=XML,frame=lines]{json1.html}

\begin{mdframed}[font=\footnotesize, frametitle={Question-Factuality Categorization}]
\begin{verbatim}
"categories": [
   {
    "name": "factoid",
    "probability": 0.25,
    "description": 
    "A question seeking a specific, concise piece of information
     or a short fact about a particular subject, such as a
     name, date, or number."
   },
   {
    "name": "non-factoid-experience",
    "probability": 0.75,
    "description":
    "A question to get advice or recommendations on a particular
     topic."
   }
]
\end{verbatim}
\end{mdframed}
\vspace{0.1cm}
%\dc{why do we need the entire json (with all parentheses etc.). Can we just show the relevant info?}
%\ym{given that we want this to be the reference pub for liverag, i think it is important we show the real JSON}
Note that the desired probabilities of occurrence of each question category in the benchmark can be explicitly defined via the attribute \texttt{probability}. Table~\ref{tab:question categories} details a set of general-purpose question categorizations and their respective categories, which can be used for most corpora.
%ym: I removed this as I wrote it in the followin subsection where it should be (and i stole the last sentence thx zohar:)
%When generating a question, the prompt will contain a single category from each of the defined categorizations, and these jointly define the type of question to be generated as detailed later. Note that by allowing all combination of categories we enable a combinatorial number of options, resulting in a highly diverse dataset.

End-user categorizations are defined analogously to question categorizations. The snippet below shows how to specify a categorization of end-users defining their expertise. We chose this categorization for end-users as our default general-purpose one since it applies as well to most corpora. \\

\begin{mdframed}[font=\footnotesize, frametitle={User-Expertise Categorization}]
\begin{verbatim}
"categories": [ 
   {
    "name": "expert",
    "probability": 0.50,
    "description": "a specialized user with deep understanding
     of the corpus."
   },
   {
    "name": "novice",
    "probability": 0.50,
    "description": "a regular user with no understanding of
     specialized terms."
   }
]
\end{verbatim}
\end{mdframed}
\vspace{0.1cm}

\begin{table*}[ht!]
\small
\centering
\caption{Question Categories. The examples in parentheses are for illustration only and are not necessarily part of the description to be used for generation.} 
\label{tab:question categories}
\begin{tabular}{c|c|p{11.5cm}} % Adjust column width as needed

\hline
\textbf{Categorization} & \textbf{Category} & \textbf{Description} \\ \hline
\multirow{4}{*}{Factuality} & \multirow{2}{*}{factoid} & question seeking a specific, concise piece of information or a short fact about a particular subject, such as a name, date, or number (e.g., {\em `When was Napoleon born?'}). \\ \cline{2-3}
 & \multirow{2}{*}{open-ended} & question inviting detailed or exploratory responses, encouraging discussion or elaboration. (e.g., {\em `what caused the French revolution?')}.\\ \hline
 
\multirow{5}{*}{Premise} & \multirow{2}{*}{direct} & question that does not contain any premise or any information about the user) (e.g., {\em `what is the fee for speeding in Italy?'}) \\ \cline{2-3}
& \multirow{3}{*}{with-premise} & question starting with a very short premise, where the user reveals their needs or some information about himself (e.g., {\em `I have an H1-B visa for the United States. Is there a limit to how many times I can exit and enter the country in a year?'}). \\ \hline

\multirow{11}{*}{Phrasing} & \multirow{3}{*}{concise-and-natural} & phrased in the way people typically speak, reflecting everyday language use, without formal or artificial structure. It is a concise direct question consisting of less than 10 words (e.g., {\em `what's the weather like in Paris now?'}). \\ \cline{2-3}
& \multirow{3}{*}{verbose-and-natural} & phrased in the way people typically speak, reflecting everyday language use, without formal or artificial structure. It is a a relatively long question consisting of more than 9 words (e.g., {\em `I thought of visiting Paris this year, not sure when is the best time. How is it like in the summer?'}).\\ \cline{2-3}
& \multirow{2}{*}{short-search-query} & phrased as a typed web query for search engines (only keywords, without punctuation and without a natural-sounding structure). It consists of less than 7 words (e.g., {\em `Paris weather August'}).\\ \cline{2-3}
& \multirow{3}{*}{long-search-query} & phrased as a typed web query for search engines (only keywords, without punctuation and without a natural-sounding structure). It consists of more than 6 words (e.g., {\em `Paris, France temperature humidity climate summer vs fall'}). \\ \hline

\multirow{6}{*}{Linguistic variation}  & \multirow{3}{*}{similar-to-document} & phrased using the same terminology and phrases appearing in the document (e.g., for the document `The Amazon River has an average discharge of about 215,000–230,000 m3/s', {\em `what is the average discharge of the Amazon river'}). \\ \cline{2-3}
 & \multirow{3}{*}{distant-from-document} & phrased using terms completely different from the ones appearing in the document (e.g., for a document `The Amazon River has an average discharge of about 215,000–230,000 m3/s', {\em `How much water run through the Amazon?'}).\\ \hline  
\end{tabular}

\end{table*}

\noindent It is possible to define additional user categorizations depending on the RAG corpus. For instance, in a healthcare RAG application, one could add {\em patient, doctor,} and {\em public health authority}, or in a RAG-based embassy chatbot {\em diplomat, student, worker,} and {\em tourist}.

\subsection{Generation Stage}
\label{sec:methodology}

%\begin{figure*}[h]
%    \centering
%    \includegraphics[width=\linewidth]{figures/datamorgana.png}
%    \caption{DataMorgana Incremental Generation}
%    \label{fig:datamorgana-architecture}
%\end{figure*}

The benchmark is built incrementally one Q\&A pair $(q_i, a_i)$ at a time.
Each pair is generated by invoking an LLM with a prompt automatically instantiated by \datamorgana according to the configuration file. Note that the structured parts of the configuration file (e.g., {\tt name, probability} in each category) are used behind the scenes to instantiate the prompt \datamorgana builds, while the {\tt description} value is inserted ``as is'' in the prompt. This gives a lot of freedom to \datamorgana users, who can iterate as needed with the description of categories when generating a benchmark\footnote{Note that in this early release of \datamorgana, we are not enforcing consistency among categories. We will wait for our early beta testers to experiment with the tool before deciding whether this is a needed capability, or whether the LLM can handle such inconsistencies on its own, e.g., via the filtering stage described below.}.  %For instance, if a user categorization covers age, and includes {\tt infant} as a category, while at the same time specifies a profession categorization with {\tt physician} as a category, it could lead to weird instructions to the LLM. We will wait for our early beta testers to experiment with the tool before deciding whether this is a needed capability, or whether the LLM can handle such inconsistencies on its own, e.g.\ via the filtering stage described below. In the meantime, we rely on users to define coherent categorizations and associated categories. 
%\dc{The discussion on limitations is too early and is not fluent with the system description. Maybe move it to the discussion on limitation at the end, (or even the appendix.}\simo{+1}

The generation process follows the steps depicted in Figure \ref{fig:datamorgana}:

\begin{enumerate}
\item  A user category $u_i$ and a question category $c_j$ are selected per categorization according to their distribution probabilities, as specified in the configuration file. So if we use the general purpose question categorizations from Table~\ref{tab:question categories} and the User Expertise Categorization detailed before this results in a combination of one user category and four question categories, $(u_1,c_1,\ldots,c_4)$. This tuple together with the natural-language description associated with each category is used to instantiate the prompt template.  Note that, by allowing all combinations of categories, we enable a combinatorial number of options, resulting in a highly diverse benchmark.
\item A document \(d_i\) is sampled from the RAG corpus and added to the prompt.
%\lle{to be completed - technical details of the sampling} \simo{In our experiments we basically substituted the stochastic sampling process with a  deterministic selection to make sure we generate a specific number of questions for each document. In the general usage of DataMorgana, especially needed if the corpus contains more documents than the QA pairs to generate the sampling process will be needed. DataMorgana implements the possibility to perform a multi-strata stratified sampling schema where the strata can be configured and depends on facet values appearing in the corpus (of course, if no strata is specified there will be a pure random sample).}
\item The chosen LLM is invoked with the instantiated prompt (a different prompt at each turn) to generate $k$ candidate question-answer pairs\footnote{We set $k$=3 in our experiments.} $(q_i, a_i)$ about $d_i$.
\item A filtering step is conducted to verify that these candidate pairs satisfy the constraints expressed in the prompt (e.g., be context-free), adhere to the categories specified by \(u_i,c_j\), and that the question answers are faithful to $d_i$. If multiple pairs satisfy the quality requirements, one is sampled.
%\lle{not sure here if filtering should be described over a single QA pairs with candidates, or over the entire benchmark} \simo{The filtering basically scores each QA candidate, so it should be described over single QA pairs.}
    
\end{enumerate}

%\ym{shouldn't we remove "for starting a conversation" in the prompt? Also  I believe the slots in the prompt eg "description of user category 1" should be replaced by the terminology above namely $c_i$ etc to ease the reading, or vice versa use "question category 1" both in prompt and in the description above, same for passage $p_i$ instead of document} \zk{+1 for "for starting a conversation". Adding to the comment about $c_i$, etc that I agree with, would it make sense to have something like "In terms of <categorization> the question should be <category>:<description>"?}\simo{The category name is never used in the prompt, but only its description. In the current formulation, the term $c_i$ refers to the category name and not to its description, so in case we should replace it with "description of $c_i$". Honestly, I would leave it the prompt as is, since anyone can immediately understand it without reading section 3.2. Finally, regarding the "for starting a conversation" it is required, otherwise the LLm tends to generate a conversation instead of independent questions (basically the second question it generates is a follow on to the first generated question.}

%ym:removed [scenario] after discussing with Guy
\begin{mdframed}[font=\footnotesize, frametitle={Prompt Template}]
\label{prompt:generation}
\begin{alltt}
You are a user simulator that should generate [num_questions]
candidate questions for starting a conversation.

The [num_questions] questions must be about facts discussed in
the documents you will now receive. When generating the questions,
assume that the real users you must simulate, as well as the readers
of the questions, do not have access to these documents. Therefore, 
never refer to the author of the documents or the documents 
themselves. Also, assume that whoever reads the questions will read 
each question independently. The [num_questions] questions must 
be diverse and different from each other. Return only the questions 
without any preamble. Write each pair in a new line, in the following 
JSON format: '\{"question": <question>, "answer": <answer>\}.'

### The generated questions should be about facts from the
following document:
[document (d_i)]

### Each of the generated questions must reflect a user with
the following characteristics:
    - They must be [description of user category 1 (u_1)]
    - They must be [description of user category 2 (u_2)]
    \dots
### Each of the generated questions must have the following
characteristics:
    - It must be [description of question category 1 (c_1)]
    - It must be [description of question category 2 (c_2)]
    \dots
\end{alltt}
\end{mdframed}
\vspace{0.1cm}

Note that this two-stage methodology is simple and lightweight by design. 
We intentionally try to avoid approaches with a costly pre-processing stage (e.g., building a knowledge graph \cite{ragas} or performing heavy analysis on the document \cite{de2024know}) or multiple invocations for post-processing (e.g., evolving a question \cite{DeepEval})
% Unlike other approaches \ym{ which ones Zohar?}, we try to minimize costly invocations for instance. 
Also, note that we describe here only the initial features of \datamorgana that will be used for the SIGIR'2025 LiveRAG Challenge. Additional capabilities are planned to be added for additional types of benchmarks in the near future.

\label{sec:report}

\section{Experimental Settings}
\label{sec:settings}

A common way of evaluating the quality of synthetic data is via fidelity, diversity, and generalization (see \cite{alaa2022faithful} and references within). Fidelity measures the
quality of a model’s synthetic samples, and Diversity is the extent to which these samples cover the full variability of real samples. Generalization applies to processes like GANs where the generation process is based on a training set of real examples, hence does not apply to our setting.
In our setting of question generation, fidelity translates to the quality of individual questions: Each generated question should be fluent, coherent, relevant to the target application, and realistic. In other words, it should represent a plausible way a real user could interact with the system.
Diversity means that the generated questions cover all or at least many of the questions asked by humans.

In our analysis, we decided not to focus on fidelity. The reason is that recent powerful LLMs (e.g., Claude-3.5-Sonnet or GPT-4) are known to excel in generative tasks and produce high-quality text, matching and perhaps exceeding human level \cite{bubeck2023sparks}; therefore the individual question quality is typically extremely high, regardless of the specific generation strategy adopted. To further confirm this assumption, in preliminary studies, we manually annotated $\sim$200 individual questions generated by DataMorgana powered by Claude-3.5-Sonnet in terms of text quality and relevance to the document used to generate each question. We observed close to perfect results. 
We note that for Q\&A pairs, fidelity includes the quality and specifically correctness of the answer. In a preliminary analysis, we observed that the answers generated by \datamorgana are typically faithful to the original document and that the Q\&A filtering stage helps remove bad generations. However, 
our focus here is on the quality of the questions rather than answers, hence we do defer further investigation of the answer quality to future research.

Our analysis below focuses on the diversity/coverage aspect, which is still an open problem due to the tendency of LLMs to generate obvious responses to input prompts, which in our scenarios means they mostly generate specific types of questions and neglect other interaction types.

\subsection{Baselines}\label{sec:baselines}
We compare \datamorgana with the following synthetic data generation methods:
\begin{itemize}
    \item Vanilla: this strategy repeatedly uses the same exact process to generate questions from different documents, namely the LLM instructions appearing in the prompt are always the same, and the only part that varies is the input document. This baseline corresponds to the usage of \datamorgana without configuration. This is probably the most common strategy to generate synthetic benchmarks \cite{chen2024benchmarking, wang2024domainrag, wang2024omnieval, li2024enhancing}.
    \item Know Your RAG: we re-implemented the solution proposed by \citet{de2024know} described in Section~\ref{sec:related}. The original solution generates four question types: Single-fact, reasoning, summary, and unanswerable questions. We excluded the latter since, while it is fitting for reading comprehension, it is too challenging in a RAG context (especially with large RAG corpora) to guarantee that no document in the corpus can answer the question.

    \item DeepEval: we chose DeepEval~\cite{DeepEval} as a representative of unplished commercial solutions. We used its default setting that enables evolving questions with one step of evolution, where the type is drawn uniformly at random from the 7 possible evolutions. We chose DeepEval since it is quite adopted, their git repo has 4.3K stars and 358 forks, and their data generation code is easy to run and flexible enough to allow generating multiple questions per document (required for the experiments below).

\end{itemize}

For a fair comparison, all tested generation methods leverage \texttt{Claude-3.5 Sonnet v2}\footnote{\url{https://www.anthropic.com/claude/sonnet}} with default parameters as LLM backbone. 

\subsection{Experiments Corpora}
To demonstrate the capabilities of \datamorgana, we generated synthetic data from two different corpora:
\begin{itemize}
    \item COVID-19 Open Research Dataset (CORD-19) \cite{de2024know}: this corpus contains scientific papers on COVID-19 and related historical coronavirus research. To allow us to compare with human-generated questions, we selected the 147 articles that biomedical experts used when generating the 2019 questions appearing in the COVID-QA dataset \cite{moller-etal-2020-covid}. This healthcare scenario serves to show how \datamorgana can be easily configured to adapt to a domain-specific corpus. 
    \item Wikipedia: Wikipedia is a free online encyclopedia that contains millions of articles about general human knowledge. To allow us to compare with human-generated questions, we considered the real user questions along with the Wikipedia passages containing their answer as they appear in the NQ dataset \cite{kwiatkowski-etal-2019-natural}. More specifically, we used the 2889 questions appearing in the test set of the open version of NQ \cite{lee-etal-2019-latent}. This dataset serves us to show the effectiveness of \datamorgana in general-purpose scenarios.
    % we selected the 2889 passages from Wikipedia English articles  %\zk{Is it possible to reconstruct this corpus of 2682 articles? Is that published as a part of NQ? If not, maybe we should have an appendix explaining how to reconstruct this.}\simo{The articles (actually the potentially long passages containing the answer) are taken from the original NQ corpus (not the open version). There are many papers doing the same including our SIGIR paper.} \ym{so Simone can you add a sentence to  this effect, namely that following and you cite a few papers we are taking a subset from NQ?} 
    % containing the answers to the real user questions appearing in the open version of the Natural Questions dataset (NQ-open) \cite{lee-etal-2019-latent}. As in \cite{10.1145/3626772.3657834}, these passages have been obtained by articles \simo{WIP} are  
\end{itemize}

\subsection{Configuration of Question and User Categorizations}
In both scenarios, we used the question categorizations and categories as detailed in Table~\ref{tab:question categories}. Regarding the user categorizations, in the Wikipedia scenario, we employed our default (general-purpose) user categorization.
For the CORD-19 corpus, we designed a categorization specific to the healthcare scenario:

\begin{itemize}
    \item \texttt{patient}: a regular patient who uses the system to get basic health information, symptom checking, and guidance on preventive care.
    \item \texttt{medical-doctor}: a medical doctor who needs to access some advanced information.
    \item \texttt{clinical-researcher}: a clinical researcher who uses the system to access population health data, conduct initial patient surveys, track disease progression patterns, etc.
    \item \texttt{public-health-authority}: a public health authority who uses the system to manage community health information dissemination, be informed on health emergencies, etc.
\end{itemize}

\section{Assessing Diversity}
\label{sec:assessing_diversity}
\subsection{Qualitative Diversity Exploration}
\label{subsec:qualitative_diversity}

\begin{table*}[ht]
\centering
\caption{Random Sample of questions generated by different methods. Sophisticated terms are highlighted in bold; questions in search query format are underlined.}
\label{tab:examples}
\footnotesize
\begin{tabular}{c|l}
\textbf{Model} & \textbf{Random Sample of Questions} \\
\hline
%\multirow{12}{*}{Vanilla} & Based on forecasting models, what was predicted to be the average percentage increase of COVID-19 cases in China from February 19-28, 2020? \\
\multirow{10}{*}{Vanilla} & How common are \textbf{co-infections} in people who have influenza, and why is this important for treatment? \\
& How do humans typically get infected with \textbf{hantavirus}, and what activities put people at higher risk of infection? \\
& How do humans typically get infected with \textbf{pathogenic arenaviruses}? \\
& How does the protein \textbf{Prohibitin (PHB)} affect the life cycle of the \textbf{lymphocytic choriomeningitis} virus? \\
& What are the current limitations of seasonal influenza vaccines that make them less effective than desired? \\
& What are the main approaches being explored for developing a universal influenza vaccine using viral vectors? \\
& What are the main clinical symptoms and warning signs of severe \textbf{adenovirus type 55} infection in otherwise healthy adults? \\
%& What is the 'information economy paradox' in viruses and how does it affect their ability to develop resistance to treatments? \\
& What is the mortality rate for \textbf{MERS} and how does it compare to \textbf{SARS}? \\
%& What percentage of tuberculosis patients in the 2010 Shandong survey did not have persistent cough as a symptom, and why is this finding significant? \\
& What specific protective equipment and safety measures were required for healthcare workers conducting \textbf{CT scans} of COVID-19 patients? \\
%& What was the estimated time delay between getting infected with COVID-19 and showing first symptoms during the early stages of the pandemic? \\
%& What was the success rate of infection control measures in protecting radiology staff during COVID-19 at West China Hospital? \\
& What were the main routes of transmission for \textbf{SARS-CoV-2} in the early stage of the outbreak in Wuhan, and which one was more significant? \\
\hline
\multirow{10}{*}{Know Your RAG} & By how much did \textbf{pneumonia} deaths in children decrease between 2000-2013 due to new vaccines? \\
%& How do bat cells' antiviral defenses enable them to maintain persistent viral infections without dying? \\
& How do \textbf{virus-vectored} flu vaccines compare to traditional vaccines in terms of safety and immune response? \\
& How does \textbf{2-bromopalmitic acid} affect \textbf{hantavirus} host cell \textbf{mineralization patterns}? \\
& How does \textbf{EGR1} deficiency affect \textbf{BIRC5} expression during \textbf{VEEV} infection? \\
%& How effective is chest X-ray screening compared to symptom screening for detecting TB cases in European populations during COVID-19? \\
%& What evidence suggests that FPASSA-ANFIS's multi-algorithm approach performs better than single-algorithm COVID-19 forecasting models? \\
& What genetic relationship was found between \textbf{SAIBK} virus and COVID-19 vaccine strains during the early pandemic studies? \\
& What genetic similarities does the French \textbf{BCoV} strain share with Asian coronavirus strains? \\
%& What percentage of viral respiratory infections might be missed by traditional virus isolation methods compared to PCR/ESI-MS in Taiwan? \\
& What safer alternative to live virus can be used for \textbf{arenavirus neutralization testing}? \\
& What starting material did the engineered \textbf{E. coli} platform use to generate \textbf{glucose-1-phosphate} for \textbf{UDP-sugar synthesis}? \\
& What was Germany's COVID-19 infection rate compared to other European countries during early pandemic interventions in March 2020? \\
%& What was the main type of binding interaction discovered in early COVID-19 enzyme studies in Europe? \\
& What was the mortality rate of \textbf{HCPS} cases in South America during 1993-2009? \\
\hline

\multirow{10}{*}{DeepEval} & How did World War 1's social and economic conditions make the Spanish flu pandemic more deadly, leading to over 20 million deaths? \\
& How do environmental factors like \textbf{habitat fragmentation}, climate patterns, and seasons affect \textbf{hantavirus} outbreaks and rodent populations in the Americas? \\
%& How do rapid diagnostic tests help control COVID-19 through surveillance, isolation, and facility management? \\
& How do respiratory viruses affect the airways? \\
& How do viral infections change our body's immune response and inflammation levels when symptoms get worse? \\
%& How does RT-LAMP testing make HIV detection simpler compared to traditional testing methods? \\
& How would Australian-Japanese biomedical research collaboration be different today if the \textbf{AIFII} and \textbf{ConBio} conferences had never taken place? \\
& How would scientists use \textbf{VP1 sequencing} and viral testing to identify meningitis infections if an outbreak happened today? \\
%& If we reversed the male-to-female ratio in HBoV1-positive patients from 2.54:1 to 1:2.54, would it significantly change the statistical comparison with HBoV1-negative cases? \\
& What are the average and highest percentage increases in COVID-19 cases predicted for China by \textbf{FPASSA-ANFIS}? \\
& What are the advantages and challenges of using \textbf{Ad5} as a vaccine vector, particularly regarding stability, storage, delivery, and immunity issues? \\
%& What evidence supports or contradicts the idea that bats are Ebola maintenance hosts compared to other potential host species? \\
%& What is the number of nucleotide differences between SARS-CoV-2 and bat coronavirus RaTG13? \\
%& What is the relationship between milk consumption in DC and the risk of MERS-CoV transmission from young camels to people with weak immune systems? \\
%& What was the difference between the initial coronavirus cases reported in Wuhan in December 2019 and the number of cases reported by January 24, 2020? \\
& What's the difference between \textbf{TIV}, \textbf{QIV}, and \textbf{LAIV} flu vaccines, and which one provides the best protection? \\
& Which \textbf{caspases} are activated, and at what concentrations, when \textbf{HT-29 cells} are treated with \textbf{Cu2} compared to untreated cells? \\
\hline

\multirow{10}{*}{\datamorgana} & Are there new ways to make better vaccines? \\
& \uline{death rate 1918 influenza young adults} \\
& \uline{hospital screening protocols during coronavirus early outbreak} \\
& How deadly was COVID compared to \textbf{SARS} and \textbf{MERS}? \\
& How many genetic differences are there between the human coronavirus that causes COVID-19 and its closest known relative found in bats? \\ 
%& How many total fever CT examinations were performed at West China Hospital for COVID-19 suspected cases between January 21 and March 9, 2020? \\
& I live in a tropical area. When do most flu cases happen? \\
%& Is LCMV dangerous for organ transplant patients? \\
%& overcome preexisting immunity vaccine vectors \\
& \uline{transmission rate comparison between respiratory viruses} \\
& What factors increase risk of \textbf{hantavirus} outbreaks? \\
%& What kinds of diseases can be caused by mosquitoes and ticks that affect the brain and nervous system? \\
%& What percentage of pregnant women in Japan were vaccinated against H1N1? \\
& What scientific evidence do we have to counter the claims that COVID-19 was created in a lab, so I can properly address community concerns about this? \\
& What were the main symptoms of early COVID-19 cases? \\
\hline
\hline
\multirow{10}{*}{Humans}& How does \textbf{MARS-COV} differ from \textbf{SARS-COV}? \\
& How was \textbf{HFRS} first brought to the attention of western medicine ? \\
& What animal models exist for both the asymptomatic carriage of \textbf{PUUV} and \textbf{SNV}? \\
& What can respiratory viruses cause? \\
%& What is Germany's estimated mean percentage [95\% credible interval] of total population infected as of 28th March? \\
& What is \textbf{MERS} mostly known as? \\
%& What is PPE? \\
& What is \textbf{RANBP2}? \\
%& What is presented in this study? \\
& What is the transmission of \textbf{MERS-CoV} is defined as? \\
& What reduces the \textbf{antimicrobial activities of alveolar macrophages}? \\
& What regulates the broad, but less specific, virus-cell interaction in a hepatitis B infection? \\
%& What traits should the new Director General of the WHO have? \\
& Where did \textbf{SARS-CoV-2} originate? \\
%& Where does EGR1 accumulate in the cell? \\
\hline
\end{tabular}

\end{table*}

To get an initial feeling of the diversity characterizing synthetic benchmarks\footnote{Unless explicitly mentioned, all the experiments reported in this paper are conducted so that for each document in the corpus we generate the exact same number of questions appearing in COVID-QA or NQ-open datasets. In most cases, there is a single question associated with each document but on rare occasions, this number can be much larger (i.e., up to 125 for a document in the COVID-QA dataset).}, we report in Table \ref{tab:examples} a random set of questions about different articles from the CORD-19 corpus, generated by different methods. 

The first set of questions is generated by the Vanilla approach. In this case, we do not provide detailed instructions to the LLM, which therefore is left completely free in its question generation process. The resulting questions are all in natural language, and most of them appear very specific, relatively long, and characterized by sophisticated terminology, reflecting an inherent bias of the LLM towards these types of questions. 

The second and third blocks report a random sample of questions generated by Know Your RAG and DeepEval. Similarly to the Vanilla solution, these methods do not allow control of the style of the question, and consequently it is exposed to the inherent LLM bias, which tends to generate detailed and long natural questions containing sophisticated terminology. The diversification introduced by their taxonomy can be appreciated by the presence of many comparative questions (e.g., \textit{What’s the difference between TIV, QIV, and LAIV flu vaccines, and which one provides the best protection?}). However, even if in \datamorgana we did not explicitly prompt the model to generate comparative questions, in some occasions the model generates such questions (e.g., \textit{How deadly was COVID compared to SARS and MERS?}).

The fourth set of questions, obtained with \datamorgana, exhibits larger diversity, with respect to the user and question categories used in the generation phase. For instance, the question phrasing categorization contributes to creating long and short questions, as well as questions in natural form and expressed as a web search query. Similarly, it is possible to appreciate questions having a premise (e.g., \textit{I live in a tropical area. When do most flu cases happen?}), basic questions having a more simplistic terminology, typical of patients (e.g., \textit{Are there new ways to make better vaccines?}), questions from public authorities (e.g., \textit{What scientific evidence do we have to counter the claims that COVID-19 was created in a lab, so I can properly address community concerns about this?} or questions which we can expect from researchers (e.g., \textit{How many genetic differences are there between the human coronavirus that causes COVID-19 and its closest known relative found in bats?})

\begin{table*}[ht]
\caption{Most common PoS template appearing in the generated questions for each method. The bold letter groups ({\bf WP}, {VBP}, etc) represent standard part-of-speech tags. We list the frequency of the most common (Top 1) pattern and the cumulative frequency of the three most common (Top 1-3) patterns over the CORD-19 corpus.}
\label{tab:patterns}
\footnotesize
\centering
\begin{tabular}{c|c|c|c|c}
\textbf{Model} & \textbf{Most Common Starting Pattern} & \textbf{Top 1 frequency} & \textbf{Top 1-3 frequency} & \textbf{Example Questions of Top Pattern} \\ 
\hline
\multirow{3}{*}{Vanilla}    &   \multirow{3}{*}{\textbf{WP VBP DT JJ NNS} } &  \multirow{3}{*}{181/2019 (9.0\%)}  & \multirow{3}{*}{321/2019 (15.9\%)} & What are the main symptoms... \\
& & & & What are the main differences... \\
& & & & What are the typical symptoms... \\
%\cline{2-3}
%& \multirow{3}{*}{\textbf{WP VBZ DT NN NN} 75/2019 (3.7\%)} & What is the mortality rate...\\
%& & What is the detection rate... \\
%& & What is the pathogenicity rate... \\
%\cline{2-3}
%& \multirow{3}{*}{\textbf{WP VBZ DT NN IN} 65/2019 (3.2\%)} & What is the relationship between...\\
%& & What is the mechanism by... \\
%& & What is the role of... \\
\hline
\multirow{3}{*}{Know Your RAG}     &   \multirow{3}{*}{\textbf{WP VBD DT NN NN}} & \multirow{3}{*}{\textbf{56/2019 (2.8\%)}}  & \multirow{3}{*}{140/2019 (6.9\%)}  & What was the gender distribution...   \\
& & & & What was the survival rate... \\
& & & & What was the detection rate... \\
% \cline{2-3}
% & \multirow{3}{*}{\textbf{WP VBZ DT JJ NN} 50/2019 (2.5\%)} & What explains the persistent gap...   \\
% & & What is the genome length... \\
% & & What is the genomic size... \\
% \cline{2-3}
% & \multirow{3}{*}{\textbf{WP VBZ DT NN NN} 34/2019 (1.7\%)} & What is the survival rate...   \\
% & & What is the mortality rate... \\
% & & What is the cytotoxicity level... \\
\hline

\multirow{3}{*}{DeepEval}     &   \multirow{3}{*}{\textbf{WP VBD DT NNS IN} } & \multirow{3}{*}{99/2019 (4.9\%)} & \multirow{3}{*}{255/2019 (12.6\%)} & What are the effects of...   \\
& & & & What are the advantages of... \\
& & & & What are the differences in... \\
% \cline{2-3}
% & \multirow{3}{*}{\textbf{WP VBP DT JJ NNS} 94/2019 (4.7\%)} & What are the main differences..   \\
% & & What are the main symptoms... \\
% & & What are the current treatments... \\
% \cline{2-3}
% & \multirow{3}{*}{\textbf{WP VBZ DT NN IN} 62/2019 (3.1\%)} & What's the difference between...   \\
% & & What is the distribution of... \\
% & & What is the stability of... \\
\hline

\multirow{3}{*}{\datamorgana}     &   \multirow{3}{*}{\textbf{WP VBP DT JJ NNS}} & \multirow{3}{*}{\textbf{56/2019 (2.8\%)}} & \multirow{3}{*}{\textbf{118/2019 (5.8\%)}}  & What are the clinical applications...   \\
& & & & What are the typical signs... \\
& & & & What are the main types... \\
% \cline{2-3}
% & \multirow{3}{*}{\textbf{WP VBP DT JJ NN} 36/2019 (1.8\%)} & What is the average percentage...   \\
% & & What is the genomic diversity... \\
% & & What's the main reason... \\
% \cline{2-3}
% & \multirow{3}{*}{\textbf{WDT NN IN NN NNS} 26/2019 (1.3\%)} & What percentage of pneumonia deaths...   \\
% & & What percentage of healthcare workers... \\
% & & What percentage of flu infections... \\
\hline  
\hline   
\multirow{3}{*}{Humans}     &   \multirow{3}{*}{\textbf{WP VBZ DT NN IN}} & \multirow{3}{*}{200/2019 (9.9\%)} & \multirow{3}{*}{337/2019 (16.7\%)} & What is an example of...   \\
& & & & What is the difference between... \\
& & & & What is the structure of... \\
% \cline{2-3}
% & \multirow{3}{*}{\textbf{WP VBZ DT JJ NN} 73/2019 (3.6\%)} & What is a critical feature...   \\
% & & What is a key factor... \\
% & & What is the main cause... \\
% \cline{2-3}
% & \multirow{3}{*}{\textbf{WP VBD DT NN IN} 64/2019 (3.2\%)} & What was the goal of...   \\
% & & What was the effect of... \\
% & & What was the result of... \\
\hline
\end{tabular}

\end{table*}

Another observation we can derive from Table \ref{tab:examples} is that the Vanilla solution tends to repeatedly use some word expressions across multiple questions, for instance, most of the questions start with \textit{What are/is/was/were} or \textit{How do/does}, while in \datamorgana this is less frequent. To better quantify this phenomenon, as in \cite{shaib2024detectionmeasurementsyntactictemplates}, we use syntactic templates, i.e., Part-of-Speech (PoS) tag sequences, that can capture structural repetitions better than lexical patterns. In particular, we use spacy\footnote{\url{https://spacy.io/}} to extract the PoS tags of the generated questions and consider the first five PoS of each question as its syntactic template.
%ym - please check that my rephrasing reflects what you wanted to say?
%We obtain a synthetic dataset from the CORD-19 corpus using the various data generation solutions, and 
After generating a Q\&A benchmark over the CORD-19 corpus using the various solutions (baselines, \datamorgana, etc.) we discussed before,
we grouped the generated questions based on their syntactic template. In the benchmark generated with the Vanilla method, we found 573 distinct templates. This number increases to 859 and 933 with DeepEval and Know Your RAG, respectively, and gets the best result of 1248 with \datamorgana.

Table \ref{tab:patterns} reports the most frequent syntactic PoS templates appearing in the generated questions, as well as their frequency and the cumulative frequency of the three common templates. Unsurprisingly, the frequent templates are typically associated with \textit{What} questions, and this is also in line with human-generated questions, where this type of question is the most frequent. An important aspect to consider is that in the Vanilla strategy, %the most frequent template accounts for 9\% of the entire benchmark, similarly as in the human-generated COVID-QA dataset, while with \datamorgana this number is only $\sim$3\%.%
the top-3 frequent templates cumulatively account for $\sim$16\% of the entire benchmark, similarly as in the human-generated COVID-QA dataset, while with \datamorgana this number is only $\sim$6\%. 
We argue that this discrepancy should not be seen as a deficiency of \datamorgana because the human experts who generated the COVID-QA datasets were not genuine users of Q\&A systems but volunteers requested to create questions for a dataset. %, and \Nii the primary scope of \datamorgana is to generate a high-coverage dataset containing enough questions to estimate a Q\&A system performance under different conditions, and not to reproduce the exact distribution of real user traffic.
%\dc{If we take the coverage of the most 3 frequent patterns as a metric, then DM is the best (5.9),then KYR (7) then DE (12.7) then vanilla (15.9) and the worse are humans (18.7) We can extend and measure coverage for k most leading patterns k = 3,4,5,6,..}\simo{I think this is a good idea for a new metric, also we can have the template perplexity as another metric. I would save these ideas for future work.}

\subsection{Quantitative Diversity Evaluation}

\subsubsection{Diversity Metrics}
To estimate the diversity of the generated benchmark $B$, we use the following metrics, as suggested in \cite{shaib2024standardizingmeasurementtextdiversity}:

\begin{itemize}
    \item N-Gram Diversity (NDG) Score: this score represents the ratio of the unique n-gram counts to all n-gram counts in the benchmark. It is a widely used metric to compute lexical diversity. Following \citet{shaib2024standardizingmeasurementtextdiversity} we use up to $n=4$ grams.  
    More formally: \begin{equation}
        NDG(B) = \sum_{n=1}^{4}\frac{\#\text{unique n-grams in }B}{\#\text{n-grams in }B}
    \end{equation}

    \item Self-Repetition Score (SRS): this metric was introduced by \citet{salkar-etal-2022-self} and it counts the number of questions that contain at least one $n$-gram (we use $n$=4) that also appears in another question in the benchmark. We define the repetition score for a dataset as the number of questions containing repeating n-grams divided by the total number of questions in that benchmark.

    \item{Compress Ratio (CR)}: The compression ratio is the ratio between the size of the file of the benchmark, to the size of its compressed file, using \texttt{gzip}. High compression ratios imply more redundancy, i.e., less diversity:
    \begin{equation}
        CR(B) = \frac{\#\text{size of }B}{\#\text{size of compressed }B}
    \end{equation}

    We refer to this metric as \texttt{word-CR}, when applied to the file containing the generated questions, and it measures the lexical diversity of the benchmark. Conversely, we use \texttt{PoS-CR} to refer to the same metric applied to the Part-of-Speech tag sequence of the questions. In this case, the metric provides an estimate of the syntactic diversity of the benchmark.
    
    \item Homogenization Score (HS): this score computes the average similarity between all question pairs in the benchmark, more formally:

    \begin{equation}
        HS(B) = \frac{1}{|B|(|B|-1)}\sum_{q,q' \in B | q \ne q'}sim(q, q')
    \end{equation}

    Where $sim$ is a similarity function between two questions, %We refer to rouge-HS as the homogenization score using the Rouge-L similarity score \cite{lin-och-2004-automatic}. This metric is useful for computing lexical diversity. As an alternative similarity metric between two questions 
    which we compute by using the cosine similarity of the question embeddings obtained by running the \texttt{all-MiniLM-L6-v2} sentence encoder from the Sentence Transformer package\footnote{\url{https://www.sbert.net/}}. We call embeddings-HS the resulting homogenization score that we use to compute semantic diversity.
    
\end{itemize}

As discussed in \cite{shaib2024standardizingmeasurementtextdiversity} these metrics often do not correlate, since they capture different diversity dimensions.

\subsubsection{Experimental Results}

\begin{table*}[ht]
\caption{Diversity scores of the COVID-QA dataset and different synthetic datasets containing the same number of questions associated with each of the 147 clinical articles appearing in the COVID-QA dataset. In bold are the best diversity scores among the synthetic datasets (excluding ablation studies). }
\label{tab:covidQA_res}
\centering
\begin{tabular}{c|c|c|c|c|c}
& \multicolumn{3}{|c|}{\textbf{lexical diversity}} & \textbf{syntactic diversity} & \textbf{semantic diversity}\\
\textbf{Model} & \textbf{NGD ($\uparrow$)} & \textbf{SRS ($\downarrow$)} & \textbf{word-CR ($\downarrow$)} & \textbf{PoS-CR ($\downarrow$)} & \textbf{embeddings-HS ($\downarrow$)} \\
\hline
Vanilla & 1.517 & 0.920 & 5.576 & 7.861 & 0.301 \\
Know Your RAG & 2.358 & 0.613 & 3.879 & 6.271 & 0.265 \\
DeepEval & 2.415 & 0.644 & \textbf{3.535} & 5.885 & 0.251 \\
\datamorgana & \textbf{2.536} & \textbf{0.372} & 3.701 & \textbf{5.583} & \textbf{0.249} \\
\hline
DM w/o question cat. & 1.777 & 0.908 & 4.746 & 6.945 & 0.296 \\
DM w/o user cat. & 2.484 & 0.401 & 3.725 & 5.648 & 0.247 \\
\hline
\hline
Humans & 2.484 & 0.365 & 3.380 & 6.212 & 0.182 \\
\end{tabular}
\end{table*}

% \begin{table*}
% \small
% \centering
% \begin{tabular}{c|c|c|c|c|c|c}
% & \multicolumn{4}{|c|}{lexical diversity} & syntactic diversity & semantic diversity\\
% Model & NGD ($\uparrow$) & rouge-HS ($\downarrow$) & SRS ($\downarrow$) & word-CR ($\downarrow$) & PoS-CR ($\downarrow$) & embeddings-HS ($\downarrow$) \\
% \hline
% Vanilla & 2.66 & 0.11 & 0.53 & 2.67 & 5.82 & 0.07 \\
% w/o question categorizations & 2.72 & 0.11 & 0.53 & 2.66 & 5.83 & 0.06 \\
% w/o user categorizations & 3.00 & 0.04 & 0.14 & 2.51 & 5.39 & 0.05 \\
% \datamorgana & 3.02 & 0.04 & 0.14 & 2.50 & 5.40 & 0.05 \\
% \hline
% Humans & 2.58 & 0.12 & 0.36 & 2.77 & 5.75 & 0.02 \\
% \end{tabular}
% \caption{Diversity scores of the open-NQ dataset and different synthetic datasets. The open NQ dataset contains 2889 questions from 2682 different Wikipedia Articles. All the synthetic datasets maintain the exact same number of questions associated to each article.}
% \end{table*}

\begin{table*}[ht]
\caption{Diversity scores of the open-NQ dataset and different synthetic datasets containing the same number of questions associated with each of the 2682 Wikipedia Passages appearing in the open-NQ dataset. In bold are the best diversity scores among the synthetic datasets (excluding ablation studies).}
\label{tab:nq_res}
\centering
\begin{tabular}{c|c|c|c|c|c}
& \multicolumn{3}{|c|}{\textbf{lexical diversity}} & \textbf{syntactic diversity} & \textbf{semantic diversity}\\
\textbf{Model} & \textbf{NGD ($\uparrow$)} & \textbf{SRS ($\downarrow$)} & \textbf{word-CR ($\downarrow$)} & \textbf{PoS-CR ($\downarrow$)} & \textbf{embeddings-HS ($\downarrow$)} \\
\hline
Vanilla & 2.662 & 0.533 & 2.665 & 5.824 & 0.068 \\
Know Your RAG & 2.981 & 0.144 & 2.488 & 5.864 & 0.074 \\
DeepEval & 2.879 & 0.371 & \textbf{2.477} & 5.631 & 0.067 \\
\datamorgana & \textbf{3.016} & \textbf{0.140} & 2.502 & \textbf{5.397} & \textbf{0.052} \\
\hline
DM w/o question cat. & 2.722 & 0.529 & 2.662 & 5.832 & 0.064 \\
DM w/o user cat. & 2.999 & 0.138 & 2.509 & 5.394 & 0.053 \\
\hline
\hline
Humans & 2.585 & 0.357 & 2.775 & 5.753 & 0.016 \\
\end{tabular}
\end{table*}

Tables \ref{tab:covidQA_res} and \ref{tab:nq_res} report the diversity scores of the benchmarks obtained with \datamorgana and the other methods described in Section \ref{sec:baselines}.
Furthermore, as an ablation study, we also run \datamorgana without user categorizations, namely \texttt{DM w/o user cat.} and without question categorizations, namely \texttt{DM w/o question cat.}
Finally, we also report the diversity scores of the human-generated benchmarks, as an additional reference point, but as indicated in Section \ref{subsec:qualitative_diversity}, not as a gold standard of diversity.

Overall, the Vanilla solution achieves the worst results. This confirms our hypothesis that repeatedly using the same general LLM prompt without specifying the desired question characteristics exposes to the inherent bias of the LLM towards some types of questions; this consequently results in low diversity. 
\datamorgana improves the Vanilla results for all metrics (in a statistically significant\footnote{We computed Student's t-test on bootstrapped samples and obtained p-values<0.01 in all metrics.} manner). 

\datamorgana is also generally better than Know Your RAG, especially in syntactic and semantic metrics, with differences that are statistically significant for all metrics for the COVID-QA case, and for all metrics but SRS and word-CR for the Wikipedia scenario. The gap between \datamorgana and DeepEval is less pronounced, but there are still statistically significant differences in NDG, SRS, and Pos-CR in both scenarios, and in embedding-HS in the Wikipedia questions. 

From the ablation studies it is clear that most of the diversity improvement of \datamorgana is due to the question categorizations: the contribution of user categorization is marginal in the COVID-QA case, where we use four different user categories, and negligible for the open-NQ corpus, where instead we use only two user categories. 
One of the reasons for the discrepancy between the impact of the question categorizations and the user categorizations is their respective sizes: while we use a single user categorization having at most four categories (as in the COVID-QA case), we apply four question categorizations in both corpora, resulting in a combination of 32 different joint question categories. In addition, the impact of user categorization is likely to vary with the RAG application, depending on the homogeneity of users, while question categorization can be applied to any corpus and is likely to remain significant regardless of the specific RAG application. 

Another aspect we need to consider while interpreting the diversity scores is that most of the metrics, especially the lexical ones, tend to favor questions using sophisticated terms; the reason is that these terms are typically extremely specific and as such appear very few times in the generated benchmark. On the opposite, questions using simpler terminology are penalized since they contain fewer distinct words and phrases. 
As we can notice in Table \ref{tab:examples}, most of the questions generated by the Vanilla approach contain very sophisticated terminology, reflecting the LLM bias towards their utilization. \datamorgana, by using user categories such as \texttt{patient} in the COVID-QA case, or \texttt{non-expert} in the open-NQ case, mitigates this bias and produces a nice mixture of simple and sophisticated questions. Unfortunately, the metrics we use do not capture this type of diversity and actually penalize it\footnote{To verify this, we compared the diversity of a benchmark consisting of only questions from simulated expert users with a benchmark containing only questions from simulated non-expert users. The benchmark from expert users resulted more diverse in all adopted metrics.}. Therefore, we believe there is a need to explore new diversity metrics, and we leave this for future work. %\simo{I can demonstrate the above statements by reporting the diversity scores of the expert vs non-expert user categories. Also, I will train a classifier to distinguish between expert vs non-expert question and I will run this classifier on the questions generated by Vanilla and Know Your RAG. What I expect is that most of the questions will be classified as "expert".}

\begin{figure*}[h]
\centering
\begin{subfigure}{0.45\linewidth}
    \includegraphics[width=0.8\linewidth]{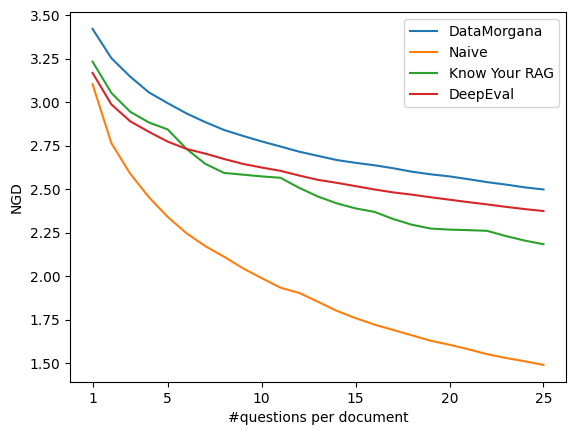}
\end{subfigure}
\hfill
\begin{subfigure}{0.45\linewidth}
    \includegraphics[width=0.8\linewidth]{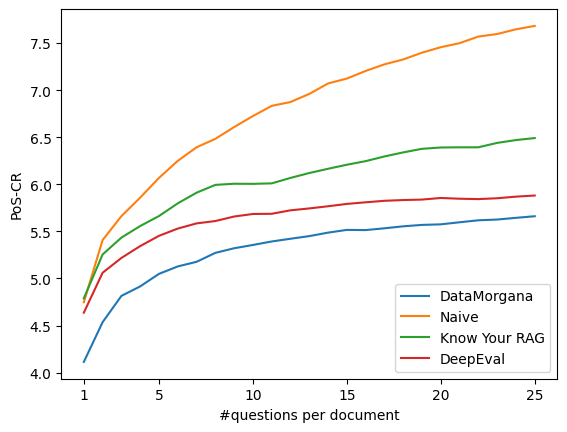}
\end{subfigure}
\caption{Lexical (NDG, the higher the better - on the left) and syntactic (PoS-CR, the lower the better - on the right) diversity of synthetic benchmarks when increasing the number of questions generated for each of the 147 documents in the COVID-QA dataset. Similar trends are observed with other metrics.}
\label{fig:increasing_questions}
\end{figure*}

As a further experiment, Figure \ref{fig:increasing_questions} reports how diversity changes when increasing the number of generated questions per document. When we generate a single question per document we are basically enforcing topical diversity, since each question is about a different document. Nevertheless, \datamorgana achieves better diversity than other solutions demonstrating that it inherently counter-balances the tendency of LLMs to generate repeated lexical or syntactic patterns. By increasing the number of questions per document, the diversity of the resulting benchmark naturally decreases, as there are multiple questions about the same topic. However, the gap between \datamorgana and the other solutions increases, demonstrating the effectiveness of the proposed method.

% \begin{figure}[h]
%     \centering
%     \includegraphics[width=\linewidth]{figures/lex_diversity_with_increased_questions.png}
%     \caption{Lexical diversity of the benchmark when increasing the number of questions generated for each of the 147 documents in the CovidQA dataset.}
%     \label{fig:lex_diversity_increasing_questions}
% \end{figure}

% \begin{figure}[h]
%     \centering
%     \includegraphics[width=\linewidth]{figures/syn_diversity_with_increased_questions.png}
%     \caption{Syntactic diversity of the benchmark when increasing the number of questions generated for each of the 147 documents in the CovidQA dataset.}
%     \label{fig:syn_diversity_increasing_questions}
% \end{figure}

% \begin{figure}[h]
%     \centering
%     \includegraphics[width=\linewidth]{figures/lex_diversity_with_increased_documents.png}
%     \caption{Lexical diversity of the benchmark when increasing the number of documents. We generate 10 questions per document.}
%     \label{fig:lex_diversity_increasing_documents}
% \end{figure}

% \begin{figure}[h]
%     \centering
%     \includegraphics[width=\linewidth]{figures/syn_diversity_with_increased_documents.png}
%     \caption{Syntactic diversity of the benchmark when increasing the number of documents. We generate 10 questions per document.}
%     \label{fig:syn_diversity_increasing_documents}
% \end{figure}

\begin{figure*}[h]
\centering
\begin{subfigure}{0.45\linewidth}
    \includegraphics[width=0.8\linewidth]{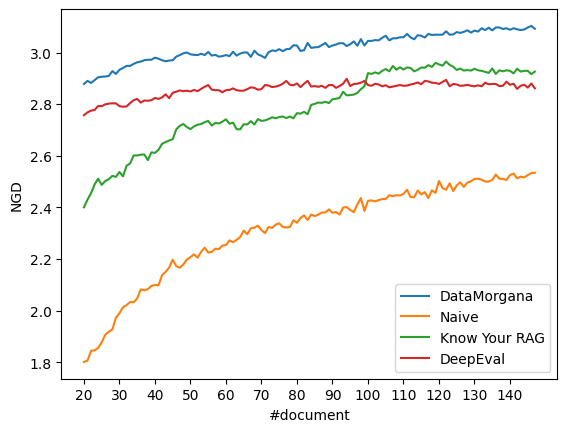}
\end{subfigure}
\hfill
\begin{subfigure}{0.45\linewidth}
    \includegraphics[width=0.8\linewidth]{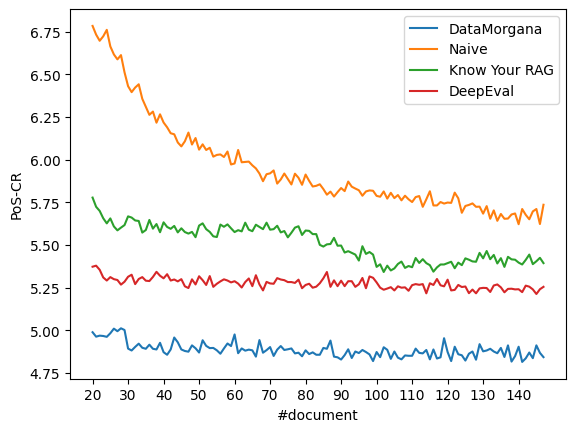}
\end{subfigure}
\caption{Lexical (NDG, the higher the better - on the left) and syntactic (PoS-CR, the lower the better - on the right) diversity of synthetic benchmarks containing 500 questions generated from an increasing number of documents in the COVID-QA dataset. Similar trends are observed with other metrics.}
\label{fig:increasing_documents}
\end{figure*}

Similar observations can be derived from Figure \ref{fig:increasing_documents}. In this case, we fix to 500 the total number of questions in the benchmark, and we increase the number of documents used to generate them, from 20 (which means that we generate 25 questions per document) to 147 (where we generate at most four questions per document). Using more documents allows for more topical diversity, and this justifies the diversity increment captured by lexical metrics. However, \datamorgana guarantees very high diversity regardless of the number of used documents and it is consistently better than other methods. %\lle{missing closing statement for the section that can be reused in intro}

\section{Conclusion}
%\ym{Explain here that while we focus on  RAG DataMorgana can be used for other types of applications that need synthetic benchmarks. Say that conversation generation and their associated diversity metrics are reserved for future work. }

In this work, we introduce \datamorgana, a benchmark generation tool that offers simple, yet rich configuration capabilities to tailor synthetic benchmarks to the expected traffic of a RAG application. %\lle{doesn't it contradict the fact that we state we are not aiming to mimic users usage?}\simo{What about "to cover the different nuances expected in real traffic of a RAG application"?}. 

\datamorgana takes as input a JSON configuration file that abstractly describes, via a semi-structured categorization representation, the expected questions and end users of the RAG application. It then automatically builds the appropriate prompts to be fed to an LLM in order to generate synthetic questions while providing good coverage for questions and users according to the configuration file. 

Through qualitative and quantitative analyses, we demonstrated that the questions generated by \datamorgana are significantly more diverse than those generated by other related question generation tools or approaches, which typically leave the choice of question type to the LLM or use internal mechanisms for controlling question diversity.

While \datamorgana was originally planned for RAG systems evaluation, it can support any application that might benefit from a high-quality and diverse Q\&A benchmark. We intend to introduce in the near future additional capabilities for generating other types of benchmarks, such as 
synthetic conversations, as well as additional controlling mechanisms, such as document sampling. \datamorgana will be made available to selected teams in the research community, as first beta testers, in the context of the upcoming SIGIR 2025 LiveRAG challenge\footnote{\url{https://sigir2025.dei.unipd.it/}}, before releasing it more widely.

\bibliographystyle{ACM-Reference-Format}
\bibliography{main}

\end{document}